\documentclass[conference]{IEEEtran}
\ifCLASSINFOpdf
\else
\fi

\usepackage{amsmath,amssymb,amsfonts}
\usepackage{algorithmic}
\usepackage{graphicx}
\usepackage{textcomp}
\usepackage{booktabs}
\usepackage{multirow}
\usepackage{balance}
\usepackage{hyperref}
\usepackage{float}

\usepackage{soul}
\usepackage[table,xcdraw]{xcolor}
\usepackage[backend=biber,style=ieee,natbib=true,sorting=nyt]{biblatex}
\DeclareUnicodeCharacter{0301}{\'{e}}

\def\BibTeX{{\rm B\kern-.05em{\sc i\kern-.025em b}\kern-.08em
    T\kern-.1667em\lower.7ex\hbox{E}\kern-.125emX}}

\usepackage{svg}
\usepackage{epstopdf}
\begin{document}
%
% paper title
% Titles are generally capitalized except for words such as a, an, and, as,
% at, but, by, for, in, nor, of, on, or, the, to and up, which are usually
% not capitalized unless they are the first or last word of the title.
% Linebreaks \\ can be used within to get better formatting as desired.
% Do not put math or special symbols in the title.
\title{Mitigating Negative Transfer with Task Awareness for Sexism, Hate Speech, and\\Toxic Language Detection}

% author names and affiliations
% use a multiple column layout for up to three different
% affiliations
% \author{\IEEEauthorblockN{Michael Shell}
% \IEEEauthorblockA{School of Electrical and\\Computer Engineering\\
% Georgia Institute of Technology\\
% Atlanta, Georgia 30332--0250\\
% Email: http://www.michaelshell.org/contact.html}
% \and
% \IEEEauthorblockN{Homer Simpson}
% \IEEEauthorblockA{Twentieth Century Fox\\
% Springfield, USA\\
% Email: homer@thesimpsons.com}
% \and
% \IEEEauthorblockN{James Kirk\\ and Montgomery Scott}
% \IEEEauthorblockA{Starfleet Academy\\
% San Francisco, California 96678--2391\\
% Telephone: (800) 555--1212\\
% Fax: (888) 555--1212}}

\author{
\IEEEauthorblockN{Angel Felipe Magnossão de Paula\IEEEauthorrefmark{1},
Paolo Rosso\IEEEauthorrefmark{1} and 
Damiano Spina\IEEEauthorrefmark{2}}
\IEEEauthorblockA{\IEEEauthorrefmark{1}Department of Computer Systems and Computation, 
Universitat Politècnica de València, València, Spain 46022 \\}
\IEEEauthorblockA{\IEEEauthorrefmark{2}School of Computing Technologies, 
RMIT University, Melbourne, Australia 3000 \\}
\IEEEauthorblockA{Email: \{adepau@doctor, prosso@dsic\}.upv.es, damiano.spina@rmit.edu.au}}

% use for special paper notices
%\IEEEspecialpapernotice{(Invited Paper)}

% make the title area
\maketitle
\balance

% As a general rule, do not put math, special symbols or citations in the abstract
\begin{abstract}
This paper proposes a novelty approach to mitigate the negative transfer problem. In the field of machine learning, the common strategy is to apply the Single-Task Learning approach in order to train a supervised model to solve a specific task. Training a robust model requires a lot of data and a significant amount of computational resources, making this solution unfeasible in cases where data are unavailable or expensive to gather. Therefore another solution, based on the sharing of information between tasks, has been developed: Multi-task Learning (MTL). Despite the recent developments regarding MTL, the problem of negative transfer has still to be solved. Negative transfer is a phenomenon that occurs when noisy information is shared between tasks, resulting in a drop in performance. This paper proposes a new approach to mitigate the negative transfer problem based on the task awareness concept. The proposed approach results in diminishing the negative transfer together with an improvement of performance over classic MTL solution. Moreover, the proposed approach has been implemented in two unified architectures to detect Sexism, Hate Speech, and Toxic Language in text comments. The proposed architectures set a new state-of-the-art both in EXIST-2021 and HatEval-2019 benchmarks.
\end{abstract}

\begin{IEEEkeywords}
Multi-task Learning, Negative Transfer, Natural Language Processing, Deep Learning 
\end{IEEEkeywords}

% For peer review papers, you can put extra information on the cover
% page as needed:
% \ifCLASSOPTIONpeerreview
% \begin{center} \bfseries EDICS Category: 3-BBND \end{center}
% \fi
%
% For peerreview papers, this IEEEtran command inserts a page break and
% creates the second title. It will be ignored for other modes.
\IEEEpeerreviewmaketitle

\section{Introduction} \label{sec:introduction}

% Machine Learning has numerous applications in fields as diverse as Natural Language Processing (NLP) (e.g., named entity recognition and hate speech detection) \cite{otter2020survey, lauriola2022introduction} or Computer Vision (CV) (e.g., object detection and object classification) \cite{voulodimos2018deep}. Generally, a single model or an ensemble of models is trained to address all the desired tasks. These models are then fine-tuned and tweaked on the chosen task until they specialize, and their performance no longer increases. Despite producing satisfactory results, a Single-Task Learning (STL) strategy ignores knowledge that may be gathered by performing other activities, allowing our model to generalize better on our original task. Furthermore, in many cases, more than the available data is needed to train a model robustly. Therefore several strategies to transfer knowledge from one task to another have been developed.

Machine Learning has numerous applications in fields as diverse as Natural Language Processing (NLP) (e.g., named entity recognition and hate speech detection) \cite{otter2020survey, lauriola2022introduction} or Computer Vision (CV) (e.g., object detection and object classification) \cite{voulodimos2018deep}. Generally, a single model or an ensemble of models is trained to address all the desired tasks. These models are then fine-tuned and tweaked on the chosen task until they specialize, and their performance no longer increases. Despite producing satisfactory results, a Single-Task Learning (STL) strategy ignores knowledge that may be gathered from datasets of related tasks, allowing our model to generalize better on our original task. Furthermore, in many cases, more than the available data is needed to train a model robustly. Therefore, several strategies to transfer knowledge from one task to another have been developed \cite{kulis2011you}.

Multi-Task Learning (MTL) \cite{Ruder2017AnOO, Zhang2021} is a new area of study that aims at exploiting the synergy between different tasks to reduce the amount of data or computational resources required for these activities. This approach aims at improving generalization by learning multiple tasks simultaneously.
The \textit{soft} \cite{wu2020aggressive,wang2022exploring} or \textit{hard parameter-sharing} \cite{fang2022,Freitas2022} strategies are two of the most commonly used 
methods for MTL employing neural networks.
In soft parameter-sharing, task-specific networks are implemented, while feature-sharing methods handle cross-task communication to encourage the parameters to be similar. Since the size of the multi-task network grows linearly with respect to the number of tasks, an issue with soft parameter-sharing systems is given by scalability.
In hard parameter-sharing, the parameter set is split into shared and task-specific operations.
It is commonly implemented with a shared encoder and numerous task-specific decoding heads \cite{Zhang2021}. One of the benefits of this method is the minimization of overfitting \cite{Ruder2017AnOO}.

Multilinear relationship networks \cite{long2017learning} enhanced this architecture by imposing tensor normal priors on the fully connected layers' parameter set. The branching sites in the network are set ad-hoc in these works, which can result in inefficient job groupings. To address this limitation, tree-based approaches \cite{lu2017fully, vandenhende2019branched} have been proposed. Despite the improvement introduced by those works, jointly learning multiple tasks might lead to \textit{negative transfer} \cite{vandenhende2020revisiting, Wu2020Understanding} 
% if useless information is shared among the tasks. During training, the hard parameter-sharing
if noisy information is shared among the tasks. During training, the hard parameter-sharing
encoder learns to construct a generic representation that focuses on extracting specific features from the input data. Nevertheless, a subset of these features may provide critical information for a given decoder head but introduces noise to another decoder to solve its respective task. Hence, negative transfer refers to situations in which the transfer of information results in a decrease in the overall model performance.

In this work, we propose a new approach to overcome the negative transfer problem based on the concept of Task Awareness (TA). This approach enables the MTL model to take advantage of the information regarding the addressed task.
The overarching goal is for the model to handle its internal weight for its own task prioritization. Unlike the State-Of-The-Art (SOTA) approaches (later presented in Section \ref{sec:related-work}), the proposed solution does not require a recursive structure, saving time and resources. Moreover, we designed two mechanisms based on the TA approach and implemented them in the creation of two Multi-Task Learning TA (MTL-TA) architectures to address SOTA challenges: Sexism, Hate Speech, and Toxic Language detection. The source code is publicly available.\footnote{\url{https://github.com/AngelFelipeMP/Mitigating-Negative-Transfer-with-TA}}

The main contributions of our work are as follows:
\begin{itemize}
\item We propose the use of the TA concept to mitigate the negative transfer problem during MTL training.
\item Design of the Task-Aware Input (TAI) mechanism to grant the MTL models with task awareness ability to mitigate negative transfer and even improve results compared with traditional MTL models. 
\item Design of the Task Embedding (TE) mechanism to give MTL models task recognition capability to diminish negative transfer and improve the results over classic MTL solutions.
\item Creation and validation of two unified architectures to detect Sexism, Hate Speech, and Toxic Language in text comments.
\item Our proposed method outperforms the SOTA on two public benchmarks for Sexism and Hate Speech detection: (i) EXIST-2021 and (ii) HatEval-2019 datasets.
\end{itemize}

%\vspace{0.2 cm}
The rest of the paper is structured as follows. Section \ref{sec:related-work} 
presents the related works of transfer learning and MTL. Section \ref{sec:proposed_approach} describes the details of our proposed method.
Section \ref{sec:experiment-setup} illustrates the experiment setup. 
Section \ref{sec:results-and-analysis} discusses and evaluates the experimental results.
Section \ref{sec:limitations} presents the limitation of our approach.
Finally, conclusions and future work are drawn in Section \ref{sec:conclusion-and-future-works}.

\section{Related Work} \label{sec:related-work}

Transfer learning is a widespread technique in machine learning based on the idea that a model created for one task can be improved by transferring information from another task \cite{weiss2016survey, pan2009survey}.
Training a model from scratch requires a large quantity of data and resources, but there are some circumstances where gathering training data is prohibitively expensive or impossible. As a result, there is the need to construct high-performance learners trained with more easily accessible data from different tasks. Transfer learning techniques allow us to improve the results of target tasks through information extracted from related tasks.
These techniques have been effectively used for a variety of machine learning applications, including NLP \cite{ruder2019transfer, wang2011heterogeneous, prettenhofer2010cross, wang2022exploring} and CV \cite{duan2012learning, kulis2011you}. The MTL framework \cite{Ruder2017AnOO, Zhang2021}, which seeks to learn many tasks at once even when they are distinct, is a closely related learning technique to transfer learning. This approach works well and can take advantage of sharing information among tasks. Still, if the tasks are not sufficiently related, it can lead to negative transfer. The problem of negative transfer consists of performance degradation caused by noisy information being shared between tasks.

To solve this issue, several approaches for balancing learning between different tasks have been proposed based on a re-weighing of the losses  (for instance, via Homoscedastic uncertainty \cite{kendall2018multi}, Gradient normalization \cite{chen2018gradnorm} and Adversarial training \cite{sinha2018gradient}) or task prioritization \cite{guo2018dynamic, zhao2018modulation, sener2018multi}.
Further recent approaches \cite{xu2018pad, zhang2018joint,zhang2019pattern} make use of the initial predictions obtained through multi-task networks to improve, once or repeatedly, each task output, overcoming a characteristic of the previously mentioned methods that computed all the task outputs for a given input at once.
Those last approaches culminate to be very time-consuming and require a lot of computational resources due to their recursive nature.

This paper proposes two unified architectures to detect Sexism, Hate Speech, and Toxic Language in text comments.
\citet{abburi2020semi} represents the first semi-supervised multi-task approach for sexism classification. The authors addressed three tasks based on labels achieved through unsupervised learning or weak labeling. The neural multi-task architecture they proposed allows shared learning across multiple tasks via common weight and a combined loss function.
The method outperforms several SOTA baselines. 
% An MTL innovative approach was proposed by \cite{wu2020aggressive} to solve Aggressive Language Detection (ALD) together with text normalization.
\citet{wu2020aggressive} proposed an MTL innovative approach to solve Aggressive Language Detection (ALD) together with text normalization.
The authors propose a shared encoder to learn the common features between the two tasks and a single encoder dedicated to learning the task-relevant features. The proposed model achieved a significant improvement in performance concerning the ALD task.

Those last approaches inspired the mechanism we propose in this paper. The main commonality is to have additional mechanisms added to the MTL models to improve the representation sent to the task heads.
The main difference with respect to the TA approach we propose is that we enrich the model with the ability to discover by itself which task it will perform. It allows the MTL-TA models to create a suitable representation for each task head. In addition, the MTL-TA modes do not need to learn an auxiliary task, resulting in more efficiency. In fact, the TA approach allows the MTL models, at each step, to try to optimize over the task at hand. The key idea is to learn a task-relevant latent representation of the data, efficiently solving many NLP tasks \cite{wang2022exploring, 9414703}.  The resulting mechanisms are proposed in the following section.

\section{Proposed Approach} \label{sec:proposed_approach}

This section describes the details of the MTL-TA models. We first introduce the notion of TA and explain how it can be beneficial in diminishing the negative transfer \cite{vandenhende2020revisiting, Wu2020Understanding} for multi-task joint training \cite{Ruder2017AnOO}. Secondly, two different TA mechanisms are proposed in order to incorporate the task self-awareness capability into MTL models. 

The mainstream approach to supervised multi-task is the hard parameter-sharing method \cite{Zhang2021}. The model is composed of an encoder and $N$ decoders or task heads, where $N$ corresponds to the number of tasks the model is simultaneously trained \cite{WORSHAM2020120}. During execution, the encoder receives input and creates a task-agnostic latent representation that is sent to a certain task head, which is in charge of producing the final prediction.

The lack of a closer relationship between the latent representation generated by the encoder and the tasks degrades the overall MTL model performance \cite{vandenhende2020revisiting}. For the same input, the optimal latent representation for task heads are likely to be different \cite{Freitas2022}. Furthermore, the encoder representation can get prone to more demanding tasks or with a larger data volume during training \cite{Ruder2017AnOO}. These model performance deteriorations are the reflex of the negative transfer phenomenon \cite{vandenhende2020revisiting, Wu2020Understanding}, where a task head receives an inaccurate input representation to solve its respective task.

We propose two TA mechanisms to mitigate negative transfer when solving multiple NLP tasks by applying the MTL approach \cite{Zhang2021}. These mechanisms tailor, depending on the specific task that is addressed, the input representation that is sent to its respective head. In addition, our proposed MTL model still takes advantage of the generalization improvements the multi-task joint training provided. Hence, the encoder and other MTL model parts located before the task heads are updated during training for every task. It should be noted that all our proposed MTL models belong to the MTL-TA class, and they follow the conventional MTL paradigm. Therefore, only the specific task head attached to the input data is considered during the task parameter updating.

\subsection{Task-Aware Input}

The first mechanism we designed to introduce task awareness into MTL models is Task-Aware Input (TAI). To compel the encoder to generate a suitable representation for each task head, we proposed to modify the MTL conventional input for NLP tasks.

\begin{figure}[ht]
\centering
\includegraphics[width=0.9\columnwidth]{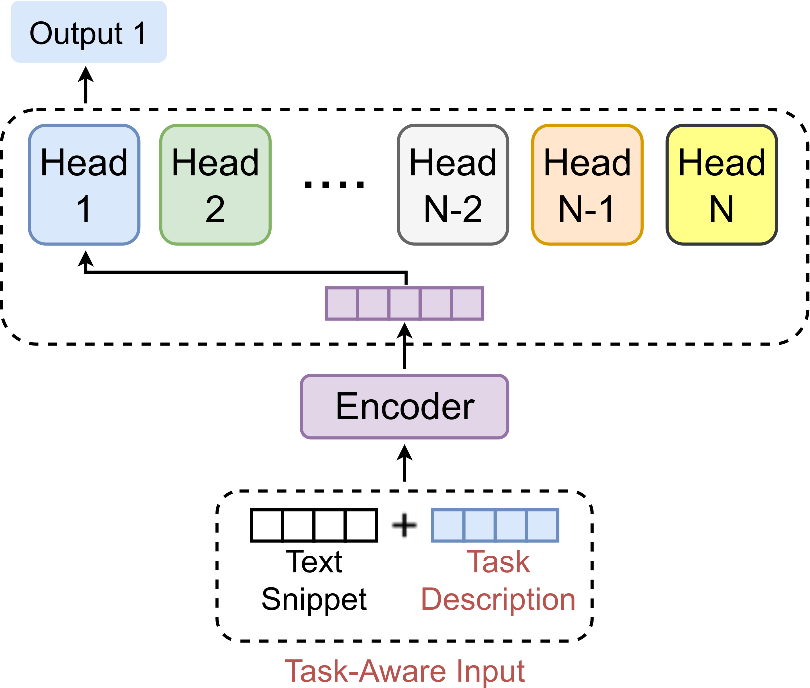}
\caption{Multi-Task Learning (MTL) model including Task-Aware Input (TAI) mechanism (MTL-TAI).}
\label{fig_mtl-tai}
\end{figure}

The TAI includes a Text Snippet (TS) plus a Task Description (TD), as shown in Fig. \ref{fig_mtl-tai}. The TS is a text chunk whose length varies according to the task. It is usually the integral input for the MTL encoders. The TD is a piece of text describing what a specific head is dealing with, such as `Sexism Detection’ and `Hate Speech Detection’. The new modified input provides context for the encoder to generate a task-centered representation. The MTL model endowed with the TAI mechanism is referred as MTL Task-Aware Input (MTL-TAI).

\subsection{Task Embedding}

\begin{figure}[ht]
\centering
\includegraphics[width=0.9\columnwidth]{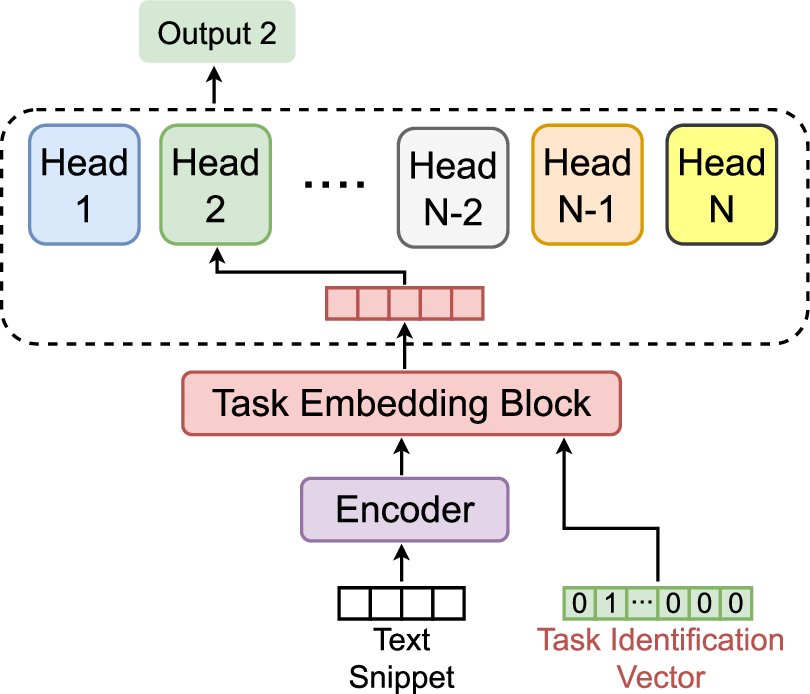}
\caption{Multi-Task Learning (MTL) model including Task Embedding (TE) mechanism (MTL-TE).}
\label{fig_mtl-teb}
\end{figure}

The second mechanism we designed to convey MTL models with the TA capability was named Task Embedding (TE). We proposed to insert an additional building block between the encoder and the task heads, which we call Task Embedding Block (TEB), as displayed in Fig. \ref{fig_mtl-teb}.  It receives two inputs: (i) the Task Identification Vector (TIV) and (ii) the latent encoder representation. The TIV is a unidimensional one-hot vector whose size is proportional to the number of task heads. Each TIV location is related to one of the task heads.

The TEB is composed of Learning Units (LU) that encompass a linear layer followed by a ReLU layer. The number of LUs is a hyperparameter that depends on the task and data, among other factors. The TEB objective is to generate a suitable representation for the task the MTL model is solving at a specific time. Hence, depending on the task, the TEB will retrieve a different output for the same exact encoder representation. It relies on the TIV to indicate for which task the TEB will generate a representation. The TIV has the number one in the location that corresponds to the task the model is about to solve. The remainder of the vector is populated with zeros, as Fig. \ref{fig_mtl-teb} reflects. 
% The MTL model equipped with the TE mechanism was named MTL Task Embedding (MTL-TE). 
The MTL model equipped with the TE mechanism is referred as MTL Task Embedding (MTL-TE).

% \begin{figure}[ht]
% \centering
% % \includegraphics[scale=0.3]{Figures/MTL-TEB.png}
% \includesvg[width=0.9\columnwidth]{Figures/MTL-TE.svg}
% \caption{Multi-Task Learning (MTL) model including Task Embedding (TE) mechanism (MTL-TE).}
% \label{fig_mtl-teb}
% \end{figure}

\section{Experimental Setup} \label{sec:experiment-setup}

% This section first describes the tasks and their respective datasets used to evaluate our approach. It then presents the implementation particularities and models for reference. Finally, we share settings for experimental details.
This section first describes the tasks and the datasets used to evaluate our approach. It then presents the implementation details and models for reference. Finally, we share the settings for the experiments.

% \ds{Can you add statistics? How many instances in train/test? Class distribution, etc. Also, indicate explicitly the evaluation campaigns/shared tasks on which has been used, and how many official runs they received. It will help to justify the choice of your experimental setup.}
% \newline
\begin{table}[H]
\caption{EXIST-2021 data distribution}
\label{tab:exist-2021}
\centering
\renewcommand{\arraystretch}{1.1}
\begin{tabular}{@{}lcccccc@{}}
%\begin{tabular}{lcccccc}
\toprule \toprule
 & \multicolumn{2}{c}{\textbf{Training}} & \multicolumn{4}{c}{\textbf{Test}}  \\ \cmidrule(lr){2-3} \cmidrule(l){4-7} 
 & Spanish       & English      & \multicolumn{2}{c}{Spanish} & \multicolumn{2}{c}{English}\\ \cmidrule(lr){2-3} \cmidrule(l){4-7} 
           & Twitter & Twitter & Twitter & Gab & Twitter & Gab \\ \cmidrule(lr){2-3} \cmidrule(l){4-7}
Sexist     & 1,741    & 1,636    & 858     & 265 & 858     & 300  \\
Not-Sexist & 1,800    & 1,800    & 812     & 225 & 858     & 192  \\ \bottomrule \bottomrule
\end{tabular}
\end{table}

\begin{table}[H]
\caption{DETOXIS-2021 data distribution}
\label{tab:detoxis-2021}
\centering
\renewcommand{\arraystretch}{1.1}
\begin{tabular}{@{}lcccc@{}}
% \begin{tabular}{lcccc}
\toprule \toprule
& \multicolumn{2}{c}{\textbf{Training}} & \multicolumn{2}{c}{\textbf{Test}}  \\ \cmidrule(lr){2-3}\cmidrule(l){4-5}
Toxic     & \multicolumn{2}{c}{1,147} & \multicolumn{2}{c}{239}  \\
Not-Toxic & \multicolumn{2}{c}{2,316} & \multicolumn{2}{c}{652}\\ \bottomrule \bottomrule
\end{tabular}
\end{table}
%
% \begin{table}[]
\begin{table}[H]
\centering
\caption{HatEval-2019 data distribution}
\label{tab:hateval-2019}
% \begin{tabular}{lcccc}
\begin{tabular}{@{}lcccccc@{}}
\toprule \toprule
& \multicolumn{2}{c}{\textbf{Training}} & \multicolumn{2}{c}{\textbf{Development}} & \multicolumn{2}{c}{\textbf{Test}}  \\ \cmidrule(lr){2-3}\cmidrule(l){4-5}\cmidrule(l){6-7}
& Spanish & English & Spanish & English  & Spanish & English \\ \cmidrule(lr){2-3}\cmidrule(l){4-5}\cmidrule(l){6-7}
Hate     & 1,741    & 1,636 & 1,741    & 1,636 & 858     & 300 \\
Not-Hate & 1,800    & 1,800 & 1,800    & 1,800 & 812     & 192 \\
\bottomrule \bottomrule
\end{tabular}
\end{table}

\subsection{Data} \label{subsac:data}
Our approach for selecting the datasets for Sexism, Hate Speech, and Toxic Language detection was based on two requirements: (i) being publicly available; (ii) having been used to evaluate a high number of ML models. We use three datasets -- EXIST-2021 \cite{rodriguez2021overview}, DETOXIS-2021 \cite{taule2021overview}, and HateEval-2019 \cite{basile-etal-2019-semeval} -- which we describe below.
\newline

\begingroup
\raggedright 
\textbf{EXIST-2021} \cite{rodriguez2021overview}: The dataset was created for the sExism Identification in Social neTworks (EXIST) shared task at Iberian Languages Evaluation Forum (IberLEF) 2021. The dataset consists of 11345 annotated social media text posts in English and Spanish from Twitter and Gab.com (Gab), an uncensored social media platform. The dataset development was supervised and monitored by experts in gender issues. The EXIST was the first challenge on Sexism detection in social media, whose objective was to identify sexism in a wide sense, from explicit misogyny to more implicit sexist behaviors. The challenge received 70 official runs for the Sexism identification task. It is a binary classification where the samples belong to the Sexist class or the Not-Sexist class. The official evaluation metric was accuracy, and data was split into training and test sets. Table \ref{tab:exist-2021} shows the data distribution.
\endgroup
\newline

\begingroup
\raggedright 
\textbf{DETOXIS-2021} \cite{taule2021overview}: The dataset was collected for the DEtection of TOxicity in comments In Spanish (DETOXIS) shared task at IberLEF 2021. The objective of the shared task was toxic language detection in comments to various online news articles regarding immigration. The proposed annotation methodology focused on diminishing the subjectivity of toxicity labeling considering contextual information (e.g., linguistic features and conversational threads). The team that worked on the data annotation was composed of trained annotators and expert linguists. The dataset consists of 4354 text comments from Twitter in Spanish and provides labels for Toxic Language detection. The task is characterized as a binary classification where the samples are divided between the Toxic and Not-Toxic classes. More than 30  teams evaluated their machine learning model in the collected dataset in the participation for DETOXIS shared task. The official data evaluation metric was F1-score in the Toxic class, and the data were divided into training and test sets. Table \ref{tab:detoxis-2021} shows the data distribution.
\endgroup
\newline

\begingroup
\raggedright 
\textbf{HatEval-2019} \cite{basile-etal-2019-semeval}: The dataset was constructed for the Detection of Hate Speech Against Immigrants and Women in Twitter (HatEval) shared task, which was part of the International Workshop on Semantic Evaluation (SemEval) 2019. The dataset comprises 19600 tweets published in English and Spanish and supplies labels for Hate Speech detection. The data collection methodology employed different gathering strategies: (i) monitoring likely victims of hate accounts; (ii) downloading the records of recognized haters; (iii) filtering Twitter streams with keywords. The annotation was performed by experts and crowdsourced contributors tested for reliable annotation. The task was defined as a binary classification where the samples are associated with the Hateful class or the Not-Hateful class. The data is composed of training, development, and test sets, and the official evaluation metric was the F1-macro, which is the unweighted mean of the F1-score calculated for the two classes. HatEval was one of the most popular shared tasks in SemEval 2019, with more than 100 submitted runs for Hate Speech detection. We can see the dataset distribution in Table \ref{tab:hateval-2019}.
\endgroup

\subsection{Implementation Details}

The encoder was constructed using a popular BERT \cite{devlin-etal-2019-bert} version for Spanish called BETO \cite{canete2020spanish}, followed by max and mean pooling calculation over its output. BETO has 12 self-attention layers, each with 12 attention-heads, using 768 as the hidden size with around 110 million parameters. 
BETO receives a text sequence and returns a hidden representation dimensionally equivalent to its hidden size for each token that belongs to the sequence. The latent encoder representation is created by a concatenation of max pooling and mean pooling calculation on the entire 768-dimensional sequence of tokens returned by BETO. 
Regarding the TE approach, the TEB preserves the same dimension of the latent encoder representation.

The task heads are linear classifiers whose input dimension corresponds to the latent encoder representation, and the output depends on the task. In the case of binary classification, the linear classifier returns two values, and the higher value corresponds to the predicted class.
% Furthermore, for the MTL-TE model,
Furthermore,
the TDs for the EXIST-2021 \cite{rodriguez2021overview}, DETOXIS-2021 \cite{taule2021overview}, and HatEval-2019 \cite{basile-etal-2019-semeval} datasets are, respectively, the following pieces of text: 'Sexism detection’, ‘Toxic Language detection’, and ‘Hate Speech detection’. 

The models were trained using the optimization algorithm AdamW \cite{ilya2019} with a linear decay learning rate schedule and a learning rate varying from 5e-6 to 1e-4. In the learning process, we trained our model for 15 epochs with a dropout of 0.3 and batch size of 64. Additionally, we experimented with 1 up to 3 LUs. Similar to the early stopping strategy \cite{caruana2000overfitting}, we adopted the model with the best performance within the epochs based on the task's official metric.

\subsection{Comparison Models} \label{subsec:comparison-models}

% \ds{Maybe move this section above Implementation Details?}
% \ds{They are all baselines: maybe we should use different terminology to introduce STL and MTL? Do we need to differentiate between the two categories?}
% \newline

We compare our approach with two types of models: (i) Baselines and (ii) SOTA models. The baselines are the two models that we implemented:

\begin{itemize}
\item\textbf{MTL} is the classic MTL model. It is constructed with the same architecture as the MTL-TA model (described in Section \ref{sec:proposed_approach}), but it does not include the TAI mechanism. Therefore, the MTL model receives only the TS as input. 
\item\textbf{STL} is the classic STL model. It has the same architecture as the MTL model, yet it encompasses only one task head. Hence, to compare this model type with the MTL models, it is necessary to train one model for each one of the addressed tasks.
\end{itemize}

The SOTA are the models which currently achieved the best performance on the datasets considered in our experiments:

\begin{table*}[ht!]
\caption{Results of The Cross-validation Experiment with 95\% Confidence Intervals}
\label{tab:cross-validation}
\centering
\renewcommand{\arraystretch}{1.1}
\begin{tabular}{cllccccc}
\toprule \toprule
&\multirow{2}{*}{\textbf{Model}} & \multicolumn{1}{c}{\multirow{2}{*}{\textbf{Task Heads}}} & & \textbf{EXIST-2021} & \textbf{DETOXIS-2021} & \textbf{HatEval-2019} & \\ \cmidrule(l){4-7} 
&                         & \multicolumn{1}{c}{} & & \textbf{Accuracy}  & \textbf{F1-score} & \textbf{F1-macro}  & \\ \midrule \midrule
&\multirow{3}{*}{STL}     & Sexism                                 & & 0.789 ± 0.011 & -- & -- & \\
&                         & Toxic-language                         & & -- & 0.640 ± 0.014 & -- & \\
&                         & Hate-speech                            & & -- & -- & 0.846 ± 0.009 & \\ \midrule
&\multirow{4}{*}{MTL}     & Sexism + Toxic-language                & & 0.788 ± 0.011 & 0.628 ± 0.014& -- & \\
&                         & Sexism + Hate-speech                   & & 0.791 ± 0.011 & -- & 0.843 ± 0.009 & \\
&                         & Toxic-language + Hate-speech           & & -- & 0.632 ± 0.014 & 0.841 ± 0.009 & \\
&                         & Toxic-language + Hate-speech + Sexism  & & 0.799 ± 0.010 & 0.634 ± 0.014 & 0.842 ± 0.009 & \\ \midrule
&\multirow{4}{*}{MTL-TAI} & Sexism + Toxic-language                & & 0.799 ± 0.010 & 0.649 ± 0.014 & -- & \\
&                         & Sexism + Hate-speech                   & & 0.805 ± 0.010 & -- & 0.984 ± 0.003 & \\
&                         & Toxic-language + Hate-speech           & & -- & 0.649 ± 0.014 & 0.988 ± 0.003 & \\
&                         & Toxic-language + Hate-speech + Sexism  & & 0.800 ± 0.010 & 0.650 ± 0.014 & 0.980 ± 0.003 & \\ \midrule
&\multirow{4}{*}{MTL-TE} & Sexism + Toxic-language                & & 0.797 ± 0.011 & 0.653 ± 0.014 & -- & \\
&                         & Sexism + Hate-speech                   & & \textbf{0.806} ± 0.010 & -- & \textbf{0.992} ± 0.002 & \\
&                         & Toxic-language + Hate-speech           & & -- & 0.653 ± 0.014 & 0.980 ± 0.003 & \\
&                         & Toxic-language + Hate-speech + Sexism  & & 0.801 ± 0.010 & \textbf{0.659} ± 0.014 & 0.988 ± 0.003 & \\ \bottomrule \bottomrule
\end{tabular}
\end{table*}

\begin{itemize}
\item\textbf{AI-UPV} \cite{depaula2021exist}: is a deep learning architecture based on the combination of different Transformers models \cite{vaswani2017attention}. It takes advantage of ensemble methods and, during training, applies data augmentation mechanisms. It is the SOTA for EXIST-2021 \cite{rodriguez2021overview}.

\item\textbf{SINAI} \cite{plaza2021sinai}: is a BERT base model \cite{devlin-etal-2019-bert} trained using the MTL hard parameter-sharing method. In spite of addressing five tasks and six datasets, the model was focused on Toxic Language detection, while the other tasks were used as auxiliary tasks. It is the SOTA for DETOXIS-2021 \cite{taule2021overview}.

\item\textbf{Atalaya} \cite{perez-luque-2019-atalaya}: is a model based on Support Vector Machines \cite{boser1992training}. It was trained on several representations computed from FastText \cite{bojanowski2017enriching} sentiment-oriented word vectors, such as tweet embeddings \cite{mikolov2013}, bag-of-characters \cite{bojanowski2017enriching}, and bag-of-words \cite{blizard1989multiset}. It is the SOTA for HatEval-2019 \cite{basile-etal-2019-semeval}.
\end{itemize}

\subsection{Experimental Settings}

We conducted two experiments to evaluate our TA approach for mitigating negative transfer \cite{vandenhende2020revisiting, Wu2020Understanding}, as described below.

\paragraph*{Cross-Validation Experiment} To assess whether the TAI and TE mechanisms were capable of reducing the negative transfer during MTL training, we performed a cross-validation experiment. Therefore, for each one of the datasets described in Subsection \ref{subsac:data}, we aggregate the different sets that compose the dataset in a unique set. Then, we run 5-fold cross-validation on the STL, MTL, MTL-TAI, and MTL-TE models.

\paragraph*{Official Training-Test Split} In order to compare our approach to the SOTA models \cite{depaula2021exist, plaza2021sinai, perez-luque-2019-atalaya} in the utilized datasets, we carried out an experiment using the official training-test split of the respective datasets. We trained our models with the training set or a combination of the training and development sets when the last was available. After that, we evaluated the models in the test partitions.

In both experiments, we use only the data samples in the Spanish language and evaluate the models employing the dataset's respective official metrics  (described in Section \ref{subsac:data}). We explored versions that combined two and three tasks for the MTL models. Furthermore, models whose results were the highest regarding the evaluation metrics were selected.
Finally, we applied the t-test to calculate the 95\% confidence interval for the experiments results.
% Finally, we applied the t-test to calculate the 95\% confidence interval for the two experiments.
% In both experiments, we use only the data samples in the Spanish language and evaluate the models employing the dataset's respective official metrics  (described in Section \ref{subsac:data}). Furthermore, we explored versions that combined two and three tasks for the MTL models. Finally, models whose results were the highest regarding the evaluation metrics were selected.
% We applied the t-test to calculate the 95\% confidence interval for all experiments.

\section{Results and Analysis} \label{sec:results-and-analysis}

This section presents the experiment’s results and the comparison among the evaluated models described in Section \ref{sec:experiment-setup}.

\subsection{Cross-Validation Experiment}

Table \ref{tab:cross-validation} shows the cross-validation results. It is organized into three parts in the following order: model type, model’s task heads, and model’s performance. Regarding the Baseline models (described in Section \ref{subsec:comparison-models}), results show that the MTL training approach suffered negative transfer on nearly all occasions. 
% The MTL model showed improvement over the STL model only for the sexism detection task. It occurred on two occasions when the model was trained for sexism and Hate Speech detection and when it was trained on the three tasks. 
The MTL model showed improvement over the STL model only for the Sexism detection task when the model was trained for Sexism and Hate Speech detection and when it was trained on the three tasks. 
Apart from that, the STL model achieved superior performance in the rest of the explored combinations. It probably happened because the negative transfer restrained the learning process of the MTL model on all the other occasions.

According to our results, the TA mechanisms worked well to diminish negative transfer.
The MTL-TAI model equipped with the TA mechanism and the MTL-TE model equipped with the TE mechanism on all occasions achieved superior performance than the classic MTL model, as shown in Table \ref{tab:cross-validation}. The MTL-TAI and MTL-TE models also overcame results obtained by the STL model for the three evaluated tasks. In general, the MTL-TE model performs better than the MTL-TAI model. 
% It may happen because the MTL-TE model includes the TEB, which offers a few more trainable parameters exclusively dedicated to tailoring the task head's input based on the addressed task. 

\begin{table*}[ht!]
\caption{Results of The Training-Test Experiment with 95\% Confidence Intervals}
\label{tab:train-test}
\centering
\renewcommand{\arraystretch}{1.1}
\begin{tabular}{cllccccc}
\toprule \toprule
&\multirow{2}{*}{\textbf{Model}} & \multicolumn{1}{c}{\multirow{2}{*}{\textbf{Task Heads}}} & & \textbf{EXIST-2021} & \textbf{DETOXIS-2021}  & \textbf{HatEval-2019}  & \\ \cmidrule(l){4-7} 
& & \multicolumn{1}{c}{} & & \textbf{Accuracy} & \textbf{F1-score} & \textbf{F1-macro}  & \\ \midrule \midrule
& AI-UPV \cite{depaula2021exist}  &   \multicolumn{1}{c}{--} & & 0.790 ± 0.018 & -- & -- & \\
& SINAI \cite{plaza2021sinai}   &   \multicolumn{1}{c}{--} & & -- & \textbf{0.646} ± 0.031 & -- & \\
& Atalaya \cite{perez-luque-2019-atalaya} &   \multicolumn{1}{c}{--} & & -- & -- & 0.730 ± 0.022 & \\ \midrule
&\multirow{3}{*}{STL}     & Sexism                                 & & 0.790 ± 0.017 & -- & -- & \\
&                         & Toxic-language                         & & -- & 0.620 ± 0.032 & -- & \\
&                         & Hate-speech                            & & -- & -- & 0.764 ± 0.021 & \\ \midrule
&\multirow{4}{*}{MTL}     & Sexism + Toxic-language                & & 0.776 ± 0.018 & 0.639 ± 0.032 & -- & \\
&                         & Sexism + Hate-speech                   & & 0.785 ± 0.017 & -- & 0.778 ± 0.020 & \\
&                         & Toxic-language + Hate-speech           & & -- & 0.593 ± 0.032 & 0.777 ± 0.020 & \\
&                         & Toxic-language + Hate-speech + Sexism  & & 0.775 ± 0.018 & 0.629 ± 0.032 & 0.773 ± 0.021 & \\ \midrule
&\multirow{4}{*}{MTL-TAI} & Sexism + Toxic-language                & & 0.797 ± 0.017 & 0.633 ± 0.032 & - & \\
&                         & Sexism + Hate-speech                   & & \textbf{0.809} ± 0.017& - & 0.789 ± 0.020 & \\
&                         & Toxic-language + Hate-speech           & & -- & 0.628 ± 0.032 & \textbf{0.790} ± 0.020 & \\
&                         & Toxic-language + Hate-speech + Sexism  & & 0.792 ± 0.017 & 0.629 ± 0.032 & 0.782 ± 0.020 & \\ \midrule
&\multirow{4}{*}{MTL-TE} & Sexism + Toxic-language                & & 0.804 ± 0.017& 0.626 ± 0.032 & -- & \\
&                         & Sexism + Hate-speech                   & & 0.804 ± 0.017& -- & 0.786 ± 0.020 & \\
&                         & Toxic-language + Hate-speech           & & -- & 0.623 ± 0.032 & 0.786 ± 0.020 & \\
&                         & Toxic-language + Hate-speech + Sexism  & & 0.802 ± 0.017& 0.633 ± 0.032 & 0.789 ± 0.020 & \\ \bottomrule \bottomrule
\end{tabular}
\end{table*}

\subsection{Official Training-Test Split}

Table \ref{tab:train-test}, following the same organization as Table \ref{tab:cross-validation}, presents the experiment carried out on the three datasets using their respective official training-test split. We see in Table \ref{tab:train-test}
%that the MTL training was not beneficial for the classic MTL model when it dealt with the sexism detection task. 
 that the MTL training was not beneficial for the classic MTL model when addressing the sexism detection task.
The model achieved lower accuracy compared with the STL model. We believe it was again due to the negative transfer phenomenon. Nevertheless, because of the TA mechanisms, the MTL-TA and MTL-TE models mitigated the negative transfer presented in the classic MTL training, achieving higher accuracy than the STL model and the EXIST-2021 SOTA (AI-UPV \cite{depaula2021exist}).

The MTL training improves the result for Toxic Language detection over the STL baseline for the training-test experiment. In general, the MTL, MTL-TAI, and MTL-TE models achieved similar results, meaning there were low negative transfer levels for this task during the formal MTL training.

We see in Table \ref{tab:train-test} that for the training and test experiment, the MTL training improved the result of Hate Speech detection. The MTL model obtained a higher F1-macro than the HatEval-2019 SOTA (Atalaya \cite{perez-luque-2019-atalaya}) and the STL Baseline. The MTL models with the TA mechanisms improved the results even more. They mitigate the negative transfer in the traditional MTL training, and both models achieved superior F1-macro than the conventional MTL model.

\subsection{Overall Analysis}

Analyzing Tables \ref{tab:cross-validation} and \ref{tab:train-test}, we see evidence that the STL model was a competitive baseline to compare our TA approach. Therefore, the STL models achieved close or better results than the SOTA models for the training-test experiment. The STL achieved the same results as the EXIST-2021 SOTA (AI-UPV \cite{depaula2021exist}) and comparable results to the DETOXIS-2021 SOTA (SINAI \cite{plaza2021sinai}). Furthermore, the STL obtained better results than the HatEval-2019 SOTA (Atalaya \cite{perez-luque-2019-atalaya}).

Summarizing the results of the two experiments, the MTL-TA models (MTL-TAI \& MTL-TEB) outperformed both the STL and the classic MTL models. It shows that our proposed TA approach could mitigate the negative transfer presented in the conventional MTL training.

\section{Limitations} \label{sec:limitations}
In this section, we mention the main limitations of our MTL-TA models.
First, the two models depending on a powerful encoder to achieve good performance. It could be a problem for low-resource computation systems that cannot afford to use deep learning architectures such as Transformers \cite{vaswani2017attention} for the encoder. 
Secondly, dealing with a higher number of tasks means having more task heads -- increasing the number of model parameters. Therefore, MTL-TA models will require more computational power to be fine-tuned.
Finally, we wonder if the MTL-TA models have their ability to adapt to unseen tasks (e.g., few-shot learning and instruction-based prompts) reduced due to the fine-tuning process utilizing information about the tasks.
% \ds{I wonder if there is a limitation about generalizability: if you add information about the tasks, to what extent it limits the ability of the model to adapt to unseen tasks (I'm thinking in terms of few-shot learning and instructin-based prompts of LLMs}
%\ds{I´ve moved this last one to future work: is not a limitation, is an opportunity!}
%Finally, our models are not prepared to deal with learning with disagreement \cite{exist2023overview}, where it is necessary to learn from all the labels provided by the annotators rather than the aggregated gold label. This new paradigm is gaining importance in NLP, especially for tasks where often there is not only one correct label, such as Sexism, Toxic Language, and Hate Speech detection.

\section{Conclusion and Future Work} \label{sec:conclusion-and-future-works}

We proposed the TA strategy to address the negative transfer \cite{vandenhende2020revisiting} problem during MTL training. The proposed method has been translated into two mechanisms: TAI and TE.
The TAI mechanism is the inclusion of the TD information to enrich the input of the MTL model encoder. The TE mechanism is the introduction of the TEB, an extra component that receives the representation generated by the encoder plus a TIV representation. The TD and the TIV provide information regarding the task the MTL model will perform at that precise moment. 
The objective of the TAI and TE is to enable the MTL model to construct task-dependent representations for the task heads to diminish negative transfer during MTL training and improve the MTL model performance. 
We proposed two MTL models, the MTL-TAI equipped with the TAI mechanisms and the MTL-TE that includes the TE mechanism.

Our two experiments show that the TA capability reduces negative transfer during traditional MTL training and improves performance over standard MTL solutions.
We achieved competitive results compared with SOTA for the two proposed MTL-TA models for the addressed tasks: Sexism, Hate Speech, and Toxic Language detection. 
In particular, the proposed models set a new SOTA on two public benchmarks: (i) EXIST-2021  \cite{rodriguez2021overview} and (ii) HatEval-2019 \cite{basile-etal-2019-semeval} datasets, demonstrating a general performance improvement of the proposed approach with respect to both the STL and classic MTL model. The TA mechanisms proved to be a valid approach to mitigate the negative transfer \cite{ Wu2020Understanding} problem in the MTL training.

This research demonstrated how an MTL approach equipped with TA mechanism leads to performance improvement in several NLP tasks. This approach has been demonstrated to be feasible in cases where we have a scarcity of labeled data. In future studies, it would be interesting to deepen the analyses to find out how many labeled samples or volumes of information it is worth applying MTL rather than using STL.
Further analyses regarding the enrichment of the MTL model input with low-level task supervision are worth it. In this scenario, the decoder receives all or a subgroup of the encoder's hidden representations instead of just the last one. It would be interesting to analyze the impact of different encoder representations in an MTL model.
We also plan to apply MTL with TA to other scenarios, such as sexism identification under the learning with disagreement regime \cite{exist2023overview}, where it is necessary to learn from all the labels provided by the annotators rather than the aggregated gold label. This new paradigm is gaining importance in NLP, especially for tasks where often there is not only one correct label.
Finally, we would like to research unsupervised techniques to improve the suggested models and tackle the same problems (detecting Hate Speech, Toxic Language, and Sexism). For instance, Latent Dirichlet Allocation \cite{blei2003latent}, Self-Organizing Maps \cite{Miljkovic2017}, and K-Means Clustering \cite{ezugwu2022comprehensive} could be considered.

\section*{Acknowledgments}
Angel Felipe Magnoss\~{a}o de Paula has received a mobility grant for doctoral 
students by the Universitat Polit\`{e}cnica de Val\`{e}ncia.
The  work of Paolo Rosso was in the framework of the FairTransNLP-Stereotypes research project
(PID2021-124361OB-C31) on Fairness and Transparency for equitable NLP
applications in social media: Identifying stereotypes and prejudices and
developing equitable systems, funded by MCIN/AEI/10.13039/501100011033 and
by ERDF, EU A way of making Europe. Damiano Spina is the recipient of an 
Australian Research Council DECRA Research Fellowship (DE200100064). 

% trigger a \newpage just before the given reference
% number - used to balance the columns on the last page
% adjust value as needed - may need to be readjusted if
% the document is modified later
%\IEEEtriggeratref{8}
% The "triggered" command can be changed if desired:
%\IEEEtriggercmd{\enlargethispage{-5in}}

% references section

% can use a bibliography generated by BibTeX as a .bbl file
% BibTeX documentation can be easily obtained at:
% http://mirror.ctan.org/biblio/bibtex/contrib/doc/
% The IEEEtran BibTeX style support page is at:
% http://www.michaelshell.org/tex/ieeetran/bibtex/
%\bibliographystyle{IEEEtran}
% argument is your BibTeX string definitions and bibliography database(s)
%\bibliography{IEEEabrv,../bib/paper}
%
% <OR> manually copy in the resultant .bbl file
% set second argument of \begin to the number of references
% (used to reserve space for the reference number labels box)
% \begin{thebibliography}{1}

% % \bibitem{IEEEhowto:kopka}
% % H.~Kopka and P.~W. Daly, \emph{A Guide to \LaTeX}, 3rd~ed.\hskip 1em plus
% %   0.5em minus 0.4em\relax Harlow, England: Addison-Wesley, 1999.

% \end{thebibliography}

% \bibliographystyle{IEEEtran}
% \bibliography{references}
\printbibliography

@InProceedings{exist2023overview,
hidedoi={10.1007/978-3-031-28241-6_68},
author="Plaza, Laura
and Carrillo-de-Albornoz, Jorge
and Morante, Roser
and Amig{\'o}, Enrique
and Gonzalo, Julio
and Spina, Damiano
and Rosso, Paolo",
hideeditor="Kamps, Jaap
and Goeuriot, Lorraine
and Crestani, Fabio
and Maistro, Maria
and Joho, Hideo
and Davis, Brian
and Gurrin, Cathal
and Kruschwitz, Udo
and Caputo, Annalina",
title={{Overview of EXIST 2023: sEXism Identification in Social NeTworks}},
hidebooktitle="Advances in Information Retrieval",
booktitle="Proc. ECIR",
year="2023",
publisher="Springer Nature Switzerland",
hideaddress="Cham",
pages="593--599",
hideisbn="978-3-031-28241-6"
}

@inproceedings{depaula2021exist,
    title = {{Sexism Prediction in Spanish and English Tweets Using Monolingual and Multilingual BERT and Ensemble Models}},
    author = "Magnoss{\~a}o de Paula, Angel Felipe and da Silva, Roberto Fray and Schlicht, Ipek Baris",
    hidebooktitle = {Proceedings of the Iberian Languages Evaluation Forum (IberLEF 2021) co-located with the XXXVII International Conference of the Spanish Society for Natural Language Processing (SEPLN 2021)},
    booktitle = {Proc. IberLEF'21},
    hidemonth = sep,
    year = "2021",
    hideaddress = "M{\'a}laga, Spain",
    pages = "356-373",
    volme= "2943",
}

@inproceedings{Wu2020Understanding,
  author    = {Sen Wu and
               Hongyang R. Zhang and
               Christopher R{\'{e}}},
  title     = {{Understanding and Improving Information Transfer in Multi-task Learning}},
  hidebooktitle = {8th International Conference on Learning Representations, {ICLR} 2020, Addis Ababa, Ethiopia, April 26-30},
  booktitle ="Proc. ICLR",
  year      = {2020},
  bibsource = {dblp computer science bibliography, https://dblp.org}
}

@article{vandenhende2020revisiting,
  author={Vandenhende, Simon and Georgoulis, Stamatios and Van Gansbeke, Wouter and Proesmans, Marc and Dai, Dengxin and Van Gool, Luc},
  hidejournal={IEEE Transactions on Pattern Analysis and Machine Intelligence}, 
  journal=ieeetpami,
  title={{Multi-task Learning for Dense Prediction Tasks: A Survey}}, 
  year={2022},
  volume={44},
  number={7},
  pages={3614-3633},
  hidedoi={10.1109/TPAMI.2021.3054719}}

@article{WORSHAM2020120,
title = {{Multi-task Learning for Natural Language Processing in the 2020s: Where are We Going?}},
journal = {Pattern Recognition Letters},
volume = {136},
pages = {120-126},
year = {2020},
hideissn = {0167-8655},
hidedoi = {https://hidedoi.org/10.1016/j.patrec.2020.05.031},
author = {Joseph Worsham and Jugal Kalita}
}

@ARTICLE{Zhang2021,
  author={Zhang, Yu and Yang, Qiang},
  journal={IEEE Transactions on Knowledge and Data Engineering}, 
  title={{A Survey on Multi-task Learning}}, 
  year={2022},
  volume={34},
  number={12},
  pages={5586-5609},
  hidedoi={10.1109/TKDE.2021.3070203}}

@INPROCEEDINGS{Freitas2022,
  author={de Freitas, João Machado and Berg, Sebastian and Geiger, Bernhard C. and Mücke, Manfred},
  hidebooktitle={2022 International Joint Conference on Neural Networks (JCNN)},
  booktitle="Proc. IJCNN",
  title={{Compressed Hierarchical Representations for Multi-task Learning and Task Clustering}}, 
  year={2022},
  volume={},
  number={},
  pages={01-08},
  hidedoi={10.1109/IJCNN55064.2022.9892342}}

@article{Ruder2017AnOO,
  author    = {Sebastian Ruder},
  title     = {{An Overview of Multi-Task Learning in Deep Neural Networks}},
  journal   = {CoRR},
  volume    = {abs/1706.05098},
  year      = {2017},
  hideurl   = {http://arxiv.org/abs/1706.05098},
  eprinttype = {arXiv},
  eprint    = {1706.05098},
  bibsource = {dblp computer science bibliography, https://dblp.org}
}

@article{rodriguez2021overview,
  title={{Overview of EXIST 2021: sEXism Identification in Social neTworks}},
  author={Rodr{\'\i}guez-S{\'a}nchez, Francisco and Carrillo-de-Albornoz, Jorge and Plaza, Laura and Gonzalo, Julio and Rosso, Paolo and Comet, Miriam and Donoso, Trinidad},
  journal={Procesamiento del Lenguaje Natural},
  volume={67},
  pages={195--207},
  year={2021}
}

@article{taule2021overview,
  title={{Overview of DETOXIS at IberLEF 2021: DEtection of TOxicity in comments In Spanish}},
  author={Taul{\'e}, Mariona and Ariza, Alejandro and Nofre, Montserrat and Amig{\'o}, Enrique and Rosso, Paolo},
  journal={Procesamiento del Lenguaje Natural},
  volume={67},
  pages={209--221},
  year={2021}
}

@inproceedings{basile-etal-2019-semeval,
    title = "{SemEval-2019 Task 5: Multilingual Detection of Hate Speech Against Immigrants and Women in Twitter}",
    author = "Basile, Valerio  and
      Bosco, Cristina  and
      Fersini, Elisabetta  and
      Nozza, Debora  and
      Patti, Viviana  and
      Rangel Pardo, Francisco Manuel  and
      Rosso, Paolo  and
      Sanguinetti, Manuela",
    booktitle = "Proceedings of the 13th International Workshop on Semantic Evaluation",
    year = "2019",
    hideaddress = "Minneapolis, Minnesota, USA",
    hidedoi = "10.18653/v1/S19-2007",
    pages = "54--63",
}

@inproceedings{devlin-etal-2019-bert,
    title = "{BERT: Pre-training of Deep Bidirectional Transformers for Language Understanding}",
    author = "Devlin, Jacob  and
      Chang, Ming-Wei  and
      Lee, Kenton  and
      Toutanova, Kristina",
    booktitle = "Proceedings of the 2019 Conference of the North American Chapter of the Association for Computational Linguistics: Human Language Technologies",
    year = "2019",
    hideaddress = "Minneapolis, Minnesota",
    hidedoi = "10.18653/v1/N19-1423",
    pages = "4171--4186",
}

@article{canete2020spanish,
  title={{Spanish Pre-trained Bert Model and Evaluation Data}},
  author={Canete, Jos{\'e} and 
    Chaperon, Gabriel and 
    Fuentes, Rodrigo and 
    Ho, Jou-Hui and 
    Kang, Hojin and 
    P{\'e}rez, Jorge},
  journal={Practical Machine Learning for Developing Countries (PML4DC) at Eleventh International Conference on Learning Representations (ICLR)},
  volume={2020},
  pages={1--10},
  year={2020}
}

@inproceedings{vaswani2017attention,
author = {Vaswani, Ashish and Shazeer, Noam and Parmar, Niki and Uszkoreit, Jakob and Jones, Llion and Gomez, Aidan N. and Kaiser, \L{}ukasz and Polosukhin, Illia},
title = {{Attention is All You Need}},
year = {2017},
hideisbn = {9781510860964},
hidepublisher = {Curran Associates Inc.},
hideaddress = {Red Hook, NY, USA},
booktitle = {Proceedings of the 31st International Conference on Neural Information Processing Systems},
pages = {6000–6010},
numpages = {11},
hidelocation = {Long Beach, California, USA},
hideseries = {NIPS'17}
}

@inproceedings{plaza2021sinai,
    title = "{SINAI at IberLEF-2021 DETOXIS Task: Exploring Features as Tasks in a Multi-task Learning Approach to Detecting Toxic Comments}",
    author = "Plaza-del-Arco, Flor Miriam and Molina-Gonz{\'a}lez, M Dolores and Alfonso, L",
    hidebooktitle = {Proceedings of the Iberian Languages Evaluation Forum (IberLEF 2021) co-located with the XXXVII International Conference of the Spanish Society for Natural Language Processing (SEPLN 2021)},
    booktitle = {Proc. IberLEF'21},
    hidemonth = sep,
    year = "2021",
    hideaddress = "M{\'a}laga, Spain",
    pages = "580-590",
    volme= "2943",
}

@inproceedings{perez-luque-2019-atalaya,
    title = "{Atalaya at SemEval 2019 Task 5: Robust Embeddings for Tweet Classification}",
    author = "P{\'e}rez, Juan Manuel  and
      Luque, Franco M.",
    booktitle = "Proceedings of the 13th International Workshop on Semantic Evaluation",
    hidemonth = jun,
    year = "2019",
    hideaddress = "Minneapolis, Minnesota, USA",
    publisher = "Association for Computational Linguistics",
    hidedoi = "10.18653/v1/S19-2008",
    pages = "64--69",
}

@inproceedings{boser1992training,
  title={{A Training Algorithm for Optimal Margin Classifiers}},
  author={Boser, Bernhard E and Guyon, Isabelle M and Vapnik, Vladimir N},
  booktitle={Proceedings of the Fifth Annual Workshop on Computational Learning Theory},
  pages={144--152},
  year={1992}
}

@article{bojanowski2017enriching,
  title={{Enriching Word Vectors with Subword Information}},
  author={Bojanowski, Piotr and Grave, Edouard and Joulin, Armand and Mikolov, Tomas},
  journal={Transactions of the Association for Computational Linguistics},
  volume={5},
  pages={135--146},
  year={2017},
  publisher={MIT Press}
}

@inproceedings{ilya2019,
  author    = {Ilya Loshchilov and
               Frank Hutter},
  title     = {{Decoupled Weight Decay Regularization}},
  hidebooktitle = {7th International Conference on Learning Representations, {ICLR} 2019,
               New Orleans, LA, USA, May 6-9, 2019},
  booktitle="Proc. ICLR",
  year      = {2019},
  bibsource = {dblp computer science bibliography, https://dblp.org}
}

@article{caruana2000overfitting,
  title={{Overfitting in Neural Nets: Backpropagation, Conjugate Gradient, and Early Stopping}},
  author={Caruana, Rich and Lawrence, Steve and Giles, C},
  journal={Advances in Neural Information Processing Systems},
  pages={381–387},
  volume={13},
  year={2000}
}

@inproceedings{mikolov2013,
  author    = {Tom{\'{a}}s Mikolov and
               Kai Chen and
               Greg Corrado and
               Jeffrey Dean},
  title     = {{Efficient Estimation of Word Representations in Vector Space}},
  hidebooktitle = {1st International Conference on Learning Representations, {ICLR} 2013,
               Scottsdale, Arizona, USA, May 2-4, 2013, Workshop Track Proceedings},
  booktitle="Proc. ICLR",
  year      = {2013},
  bibsource = {dblp computer science bibliography, https://dblp.org}
}

@article{blizard1989multiset,
	publisher = {Duke University Press},
	year = {1988},
	title = {{Multiset Theory}},
	pages = {36--66},
	hidedoi = {10.1305/ndjfl/1093634995},
	number = {1},
	author = {Wayne D. Blizard},
	volume = {30},
	journal = {{Notre Dame Journal of Formal Logic}}
}

@inproceedings{ruder2019transfer,
  title={{Transfer Learning in Natural Language Processing}},
  author={Ruder, Sebastian and Peters, Matthew E and Swayamdipta, Swabha and Wolf, Thomas},
  booktitle={Proceedings of the 2019 Conference of the North American Chapter of the Association for Computational Linguistics: Tutorials},
  pages={15--18},
  year={2019}
}

@inproceedings{wang2011heterogeneous,
author = {Wang, Chang and Mahadevan, Sridhar},
title = {{Heterogeneous Domain Adaptation Using Manifold Alignment}},
year = {2011},
hideisbn = {9781577355144},
hidepublisher = {AAAI Press},
booktitle = {Proceedings of the Twenty-Second International Joint Conference on Artificial Intelligence - Volume Volume Two},
pages = {1541–1546},
numpages = {6},
hidelocation = {Barcelona, Catalonia, Spain},
hideseries = {IJCAI'11}
}

@inproceedings{prettenhofer2010cross,
  title={{Cross-language Text Classification Using Structural Correspondence Learning}},
  author={Prettenhofer, Peter and Stein, Benno},
  booktitle={Proceedings of the 48th Annual Meeting of the Association for Computational Linguistics},
  pages={1118--1127},
  year={2010}
}

@inproceedings{duan2012learning,
author = {Duan, Lixin and Xu, Dong and Tsang, Ivor W.},
title = {{Learning with Augmented Features for Heterogeneous Domain Adaptation}},
year = {2012},
publisher = {Omnipress},
hideaddress = {Madison, WI, USA},
booktitle = {Proceedings of the 29th International Coference on International Conference on Machine Learning (ICML '12)},
pages = {667–674},
numpages = {8},
hidelocation = {Edinburgh, Scotland},
hideseries = {ICML'12}
}

@inproceedings{kulis2011you,
  title={{What You Saw is Not What You Get: Domain Adaptation Using Asymmetric Kernel Transforms}},
  author={Kulis, Brian and Saenko, Kate and Darrell, Trevor},
  booktitle={CVPR 2011},
  pages={1785--1792},
  year={2011},
  organization={IEEE}
}

@article{weiss2016survey,
  title={{A Survey of Transfer Learning}},
  author={Weiss, Karl and Khoshgoftaar, Taghi M and Wang, DingDing},
  journal={Journal of Big Data},
  volume={3},
  number={1},
  pages={1--40},
  year={2016},
  publisher={SpringerOpen}
}

@article{pan2009survey,
  title={{A Survey on Transfer Learning}},
  author={Pan, Sinno Jialin and Yang, Qiang},
  journal={IEEE Transactions on Knowledge and Data Engineering},
  volume={22},
  number={10},
  pages={1345--1359},
  year={2009},
  publisher={IEEE}
}

@inproceedings{long2017learning,
author = {Long, Mingsheng and Cao, Zhangjie and Wang, Jianmin and Yu, Philip S.},
title = {{Learning Multiple Tasks with Multilinear Relationship Networks}},
year = {2017},
booktitle = {Proceedings of the 31st International Conference on Neural Information Processing Systems},
pages = {1593–1602},
numpages = {10},
}

@inproceedings{lu2017fully,
  title={{Fully-adaptive Feature Sharing in Multi-task Networks with Applications in Person Attribute Classification}},
  author={Lu, Yongxi and Kumar, Abhishek and Zhai, Shuangfei and Cheng, Yu and Javidi, Tara and Feris, Rogerio},
  booktitle={Proceedings of the IEEE Conference on Computer Vision and Pattern Recognition},
  pages={5334--5343},
  year={2017}
}

@inproceedings{vandenhende2019branched,
  author    = {Simon Vandenhende and
               Stamatios Georgoulis and
               Luc Van Gool and
               Bert De Brabandere},
  title     = {{Branched Multi-task Networks: Deciding What Layers to Share}},
  booktitle = {Proceedings of the 31st British Machine Vision Conference (BMVC '20)},
  hidelocation = {Virtual  Event, UK},
  hideseries = {BMVC '20},
  publisher = {{BMVA} Press},
  year      = {2020},
  hideurl       = {https://www.bmvc2020-conference.com/assets/papers/0213.pdf},
  bibsource = {dblp computer science bibliography, https://dblp.org}
}

@inproceedings{kendall2018multi,
  title={{Multi-task Learning Using Uncertainty to Weigh Losses for Scene Geometry and Semantics}},
  author={Kendall, Alex and Gal, Yarin and Cipolla, Roberto},
  booktitle={Proceedings of the IEEE Conference on Computer Vision and Pattern Recognition},
  pages={7482--7491},
  year={2018}
}

@InProceedings{chen2018gradnorm,
  title = 	 {{GradNorm: Gradient Normalization for Adaptive Loss Balancing in Deep Multitask Networks}},
  author =       {Chen, Zhao and Badrinarayanan, Vijay and Lee, Chen-Yu and Rabinovich, Andrew},
  hidebooktitle = 	 {Proceedings of the 35th International Conference on Machine Learning},
  booktitle="Proc. ICML",
  pages = 	 {794--803},
  year = 	 {2018},
  hideeditor = 	 {Dy, Jennifer and Krause, Andreas},
  hidevolume = 	 {80},
  hideseries = 	 {Proceedings of Machine Learning Research},
  hidemonth = 	 {10--15 Jul},
  publisher =    {PMLR},
  hidepdf = 	 {http://proceedings.mlr.press/v80/chen18a/chen18a.pdf},
  hideurl = 	 {https://proceedings.mlr.press/v80/chen18a.html}
}

@misc{sinha2018gradient,
  title={Gradient Adversarial Training of Neural Networks},
  author={Sinha, Ayan Tuhinendu and Rabinovich, Andrew and Chen, Zhao and Badrinarayanan, Vijay},
  year={2021},
  hidemonth={6},
  publisher={Google Patents},
  note={US Patent App. 17/051,982}
}

@inproceedings{guo2018dynamic,
  title={{Dynamic Task Prioritization for Multitask Learning}},
  author={Guo, Michelle and Haque, Albert and Huang, De-An and Yeung, Serena and Fei-Fei, Li},
  hidebooktitle={Proceedings of the European conference on Computer Vision (ECCV)},
  booktitle="Proc. ECCV",
  pages={270--287},
  year={2018}
}

@inproceedings{zhao2018modulation,
  title={{A Modulation Module for Multi-task Learning with Applications in Image Retrieval}},
  author={Zhao, Xiangyun and Li, Haoxiang and Shen, Xiaohui and Liang, Xiaodan and Wu, Ying},
  hidebooktitle={Proceedings of the European Conference on Computer Vision (ECCV)},
  booktitle="Proc. ECCV",
  pages={401--416},
  year={2018}
}

@inproceedings{sener2018multi,
author = {Sener, Ozan and Koltun, Vladlen},
title = {{Multi-Task Learning as Multi-Objective Optimization}},
year = {2018},
hidepublisher = {Curran Associates Inc.},
hideaddress = {Red Hook, NY, USA},
booktitle = {Proceedings of the 32nd International Conference on Neural Information Processing Systems},
pages = {525–536},
numpages = {12},
hidelocation = {Montr\'{e}al, Canada},
hideseries = {NIPS'18}
}

@inproceedings{xu2018pad,
  title={{Pad-net: Multi-tasks Guided Prediction-and-distillation Network for Simultaneous Depth Estimation and Scene Parsing}},
  author={Xu, Dan and Ouyang, Wanli and Wang, Xiaogang and Sebe, Nicu},
  booktitle={Proceedings of the IEEE Conference on Computer Vision and Pattern Recognition},
  pages={675--684},
  year={2018}
}

@inproceedings{zhang2018joint,
  title={{Joint Task-recursive Learning for Semantic Segmentation and Depth Estimation}},
  author={Zhang, Zhenyu and Cui, Zhen and Xu, Chunyan and Jie, Zequn and Li, Xiang and Yang, Jian},
  hidebooktitle={Proceedings of the European Conference on Computer Vision (ECCV)},
  booktitle="Proc. ECCV",
  pages={235--251},
  year={2018}
}

@inproceedings{zhang2019pattern,
  title={{Pattern-affinitive Propagation Across Depth, Surface Normal and Semantic Segmentation}},
  author={Zhang, Zhenyu and Cui, Zhen and Xu, Chunyan and Yan, Yan and Sebe, Nicu and Yang, Jian},
  booktitle={Proceedings of the IEEE/CVF Conference on Computer Vision and Pattern Recognition},
  pages={4106--4115},
  year={2019}
}

@inproceedings{abburi2020semi,
  title={{Semi-supervised Multi-task Learning for Multi-label Fine-grained Sexism Classification}},
  author={Abburi, Harika and Parikh, Pulkit and Chhaya, Niyati and Varma, Vasudeva},
  booktitle={Proceedings of the 28th International Conference on Computational Linguistics},
  pages={5810--5820},
  year={2020}
}

@inproceedings{wu2020aggressive,
  title={{Aggressive Language Detection with Joint Text Normalization via Adversarial Multi-task Learning}},
  author={Wu, Shengqiong and Fei, Hao and Ji, Donghong},
  booktitle={CCF International Conference on Natural Language Processing and Chinese Computing},
  pages={683--696},
  year={2020},
  hideorganization={Springer}
}

@inproceedings{wang2022exploring,
  title={{Exploring Topic Supervision with BERT for Text Matching}},
  author={Wang, Yuan and Xu, Maoling and Yan, Yanling and Zhao, Tingting and Chen, Yarui and Yang, Jucheng},
  hidebooktitle={2022 International Joint Conference on Neural Networks (IJCNN)},
  booktitle="Proc. IJCNN",
  pages={1--7},
  year={2022},
  hideorganization={IEEE}
}

@INPROCEEDINGS{9414703,
  author={Indurthi, Sathish and Zaidi, Mohd Abbas and Kumar Lakumarapu, Nikhil and Lee, Beomseok and Han, Hyojung and Ahn, Seokchan and Kim, Sangha and Kim, Chanwoo and Hwang, Inchul},
  booktitle={ICASSP 2021 - 2021 IEEE International Conference on Acoustics, Speech and Signal Processing (ICASSP)}, 
  title={{Task Aware Multi-Task Learning for Speech to Text Tasks}}, 
  year={2021},
  volume={},
  number={},
  pages={7723-7727},
  hidedoi={10.1109/ICASSP39728.2021.9414703}}

@article{otter2020survey,
  title={{A Survey of the Usages of Deep Learning for Natural Language Processing}},
  author={Otter, Daniel W and Medina, Julian R and Kalita, Jugal K},
  journal={IEEE Transactions on Neural Networks and Learning Systems},
  volume={32},
  number={2},
  pages={604--624},
  year={2020},
  publisher={IEEE}
}

@article{lauriola2022introduction,
  title={{An Introduction to Deep Learning in Natural Language Processing: Models, Techniques, and Tools}},
  author={Lauriola, Ivano and Lavelli, Alberto and Aiolli, Fabio},
  journal={Neurocomputing},
  volume={470},
  pages={443--456},
  year={2022},
  publisher={Elsevier}
}

@article{voulodimos2018deep,
author = {Voulodimos, Athanasios and Doulamis, Nikolaos and Doulamis, Anastasios and Protopapadakis, Eftychios and Andina, Diego},
title = {{Deep Learning for Computer Vision: A Brief Review}},
year = {2018},
issue_date = {2018},
publisher = {Hindawi Limited},
address = {London, GBR},
volume = {2018},
hideissn = {1687-5265},
hideurl = {https://doi.org/10.1155/2018/7068349},
hidedoi = {10.1155/2018/7068349},
journal = {Computational Intelligence and Neuroscience},
hidemonth = {jan},
numpages = {13}
}

@INPROCEEDINGS{fang2022,
  author={Fang, Lingyong and Liu, Gongshen and Zhang, Ru},
  hidebooktitle={2022 International Joint Conference on Neural Networks (JCNN)},
  booktitle="Proc. IJCNN",
  title={{Sense-aware BERT and Multi-task Fine-tuning for Multimodal Sentiment Analysis}}, 
  year={2022},
  volume={},
  number={},
  pages={1-8},
  hidedoi={10.1109/IJCNN55064.2022.9892116}}

@article{blei2003latent,
  title={{Latent Dirichlet Allocation}},
  author={Blei, David M and Ng, Andrew Y and Jordan, Michael I},
  journal={{Journal of Machine Learning Research}},
  volume={3},
  number={Jan},
  pages={993--1022},
  year={2003}
}

@INPROCEEDINGS{Miljkovic2017,
  author={Miljković, Dubravko},
  booktitle={2017 40th International Convention on Information and Communication Technology, Electronics and Microelectronics (MIPRO)}, 
  title={{Brief Review of Self-organizing Maps}}, 
  year={2017},
  volume={},
  number={},
  pages={1061-1066},
  hidedoi={10.23919/MIPRO.2017.7973581}}

@article{ezugwu2022comprehensive,
  title={{A Comprehensive Survey of Clustering Algorithms: State-of-the-art Machine Learning Applications, Taxonomy, Challenges, and Future Research Prospects}},
  author={Ezugwu, Absalom E and Ikotun, Abiodun M and Oyelade, Olaide O and Abualigah, Laith and Agushaka, Jeffery O and Eke, Christopher I and Akinyelu, Andronicus A},
  journal={Engineering Applications of Artificial Intelligence},
  volume={110},
  pages={104743},
  year={2022},
  publisher={Elsevier}
}
% that's all folks
\end{document}